\documentclass[11pt]{article}
\usepackage{authblk}
\usepackage{varwidth}
\usepackage{hyperref}
\usepackage{paclic33}
\usepackage{times}
\usepackage{latexsym}
\usepackage{amsmath}
\usepackage{multirow }
\usepackage{url}

\usepackage{CJKutf8}
\usepackage{array}
\usepackage{multirow}
\usepackage{color,soul}
\usepackage{graphicx}
\usepackage{amsmath}

\title{Korean-to-Chinese Machine Translation using Chinese Character as Pivot Clue}

\author[1,2,3]{Jeonghyeok Park} 
\author[1,2,3, \thanks{\hspace{0.2cm}Corresponding author. This paper was partially supported by National Key Research and Development Program
of China (No. 2017YFB0304100) and Key Projects of National Natural Science Foundation of China (U1836222 and 61733011).
}]{Hai Zhao}
\affil[1]{Department of Computer Science and Engineering, Shanghai Jiao Tong University} 
\affil[2]{\protect\begin{varwidth}[t]{\linewidth}\protect\centering Key Laboratory of Shanghai Education Commission for Intelligent Interaction \par and Cognitive Engineering, Shanghai Jiao Tong University,  China\protect\end{varwidth}}

\affil[3]{ MoE Key Lab of Artificial Intelligence AI Institute, Shanghai Jiao Tong University}
\affil[ ]{\textit {\ 117033990011@sjtu.edu.cn, zhaohai@cs.sjtu.edu.cn}}

\date{}

\begin{document}
\maketitle
\begin{abstract}


Korean-Chinese is a low resource language pair, but Korean and Chinese have a lot in common in terms of vocabulary. Sino-Korean words, which can be converted into corresponding Chinese characters, account for more then fifty of the entire Korean vocabulary. Motivated by this, we propose a simple linguistically motivated solution to improve the performance of Korean-to-Chinese neural machine translation model by using their common vocabulary.
We adopt Chinese characters as a translation pivot by converting Sino-Korean words in Korean sentence to Chinese characters and then train machine translation model with the converted Korean sentences as source sentences. The experimental results on Korean-to-Chinese translation demonstrate that the models with the proposed method improve translation quality up to 1.5 BLEU points in comparison to the baseline models.
\end{abstract}

\section{Introduction}

Neural machine translation (NMT) using sequence-to-sequence structure has achieved remarkable performance for most language pairs \cite{Bahdanau:14,Cho:14,Sutskever:14,Luonga:15}. 
Many studies on NMT have tried to improve the translation performance by changing the structure of the network model or adding new strategies \cite{Wu and Zhao:18,Zhang:18,Xiao:19}. 
Meanwhile, there are few attempts to improve the performance of the NMT model using linguistic characteristics for several language pairs \cite{Sennrich:16}. 
On the other hand, Most of the recently proposed statistical machine translation (SMT) systems have attempted to improve translation performance by using linguistic features including part-of-speech (POS) tags \cite{Ueffing:13}, syntax \cite{Zhang:07}, semantics \cite{Rafael:11}, reordering information \cite{Zang:15,Zhang:16} and so on.

In this work, we focus on machine translation between Korean and Chinese, which have few parallel corpora but share a well-known culture heritage, the Sino-Korean words. 
Chinese loanwords used in Korean are called Sino-Korean words, and can also be written in Chinese characters which are still used by modern Chinese people.
Such a shared vocabulary makes the two languages closer despite their huge linguistic difference and provides the possibility for better machine translation. 



\begin{table*}[t!]
\begin{center}
\begin{tabular}{l|l}
\hline \textbf{Systems} & \textbf{Sentences} \\ \hline
Korean & \begin{CJK}{UTF8}{mj}\underline{명령}은 아래와 같이 \underline{반포}되었다.\end{CJK} \\
HH-Convert & \begin{CJK*}{UTF8}{gbsn}\underline{命令}\end{CJK*}\begin{CJK}{UTF8}{mj}은 아래와 같이 \end{CJK}\begin{CJK*}{UTF8}{gbsn}\underline{颁布}\end{CJK*}\begin{CJK}{UTF8}{mj}되었다.\end{CJK}\\
Chinese & \begin{CJK*}{UTF8}{gbsn}命令颁布如下。 \end{CJK*}  \\
English &  The command was promulgated as follows. \\
\hline
Korean & \begin{CJK}{UTF8}{mj}\underline{양국}은 \underline{광범}한 \underline{영역}에서의 \underline{공동} \underline{이익}을 \underline{확인}했다.\end{CJK} \\
HH-Convert & \begin{CJK*}{UTF8}{gbsn}\underline{两国}\end{CJK*}\begin{CJK}{UTF8}{mj}은 \end{CJK}\begin{CJK*}{UTF8}{gbsn}\underline{广范}\end{CJK*}\begin{CJK}{UTF8}{mj}한 \end{CJK}\begin{CJK*}{UTF8}{gbsn}\underline{领域}\end{CJK*}\begin{CJK}{UTF8}{mj}에서의 \end{CJK}\begin{CJK*}{UTF8}{gbsn}\underline{共同} \underline{利益}\end{CJK*}\begin{CJK}{UTF8}{mj}을 \end{CJK}\begin{CJK*}{UTF8}{gbsn}\underline{确认}\end{CJK*}\begin{CJK}{UTF8}{mj}했다.\end{CJK}\\
Chinese & \begin{CJK*}{UTF8}{gbsn}两国在广泛的领域确认了共同利益。 \end{CJK*}  \\
English &  The two countries have confirmed common interests in a wide range of areas. \\
\hline

\end{tabular}
\end{center}
\caption{\label{1-table} The HH-Convert is Korean sentence converted by Hangul-Hanja conversion of the Hanjaro. The underline denotes Sino-Korean word and its corresponding Chinese characters in Korean sentence and HH-Convert sentence, respectively.}
\end{table*}

Because of its long history of contact with China, Koreans have used Chinese characters as their writing system, and even after adopting Hangul(\begin{CJK}{UTF8}{mj}한글\end{CJK} in Korean) as the standard language, Chinese characters have a considerable influence in Korean vocabulary.
Currently, the writing system adopted by modern Korean is Hangul, but Chinese characters continue to be used in Korean and Chinese characters used in Korean are called "Hanja".
Korean vocabulary can be categorized into native Korean words, Sino-Korean words, and loanwords from other languages.
The Sino-Korean vocabulary refers to Korean words of Chinese origin and can be converted into corresponding Chinese characters, and considerably account for about 57\% of Korean vocabulary. 
Table~\ref{1-table} shows some sentence pairs of Korean and Chinese with the converted Sino-Korean words. In Table~\ref{1-table}, some Chinese words 
are commonly observed between the converted Korean sentence and the Chinese sentence.

In this paper, we present a novel yet straightforward method for better Korean-to-Chinese MT by exploiting the connection of Sino-Korean vocabulary. 
We convert all Sino-Korean words in Korean sentences into Chinese characters and take the converted Korean sentences as the updated source data for later MT model training. 
Our method is applied to two types of NMT models, recurrent neural network (RNN) and the Transformer, and shows significant translation performance improvement. 


\section{Related Work}

There have been studies of linguistic annotation, such as dependency label 
\cite{Wu:18,Li:18a,Li:18b}, semantic role labels
\cite{Guan:19,Li:19} and so on. 
Sennrich and Haddow \shortcite{Sennrich:16} proved that various linguistic features can be valuable for NMT. 
In this work, we focus on the linguistic connection between Korean and Chinese to improve Korean-to-Chinese NMT.

There are several studies on Korean-Chinese machine translation. For example, Kim et al. \shortcite{KimCH:02} proposed verb-pattern-based Korean-to-Chinese MT system that uses pattern-based knowledge and consistently manages linguistic peculiarities between language pairs to improve MT performance. Li et al. \shortcite{Li:09} improved the translation quality for Chinese-to-Korean SMT by using Chinese syntactic reordering for an adequate generation of Korean verbal phrases.

Since Chinese and Korean belong to entirely different language families in terms of typology and genealogy,  
many studies also tried to analyze sentence structure and word alignment of the two languages and then proposed the specific methods for their concern 
\cite{Huang:00,Kim:02,Li:08}. 
Lu et al. \shortcite{Lu:15} proposed a method of translating Korean words into Chinese using the Chinese character knowledge.

There are several attempts to exploit the connection between the source language and the target language in machine translation.
Kuang et al. \shortcite{Kuang:18} proposed methods to somewhat shorten the distance between the source and target words in NMT model, and thus strengthen their association, through a technique bridging source and target word embeddings.
For other low-resource language pairs, using pivot language to overcome the limitation of the insufficient parallel corpus has been a choice \cite{Habash:09,Zahabi:13,Ahmannia:17}.
Chu et al. \shortcite{Chu:13} bulid a Chinese character mapping table for Japanese, Traditional Chinese, and Simplified Chinese and verified the effectiveness of shared Chinese characters for Chinese–Japanese MT.
Zhao et al. \shortcite{Zhao:13} used the Chinese character, a common form of both languages, as a translation bridge in the Vietnamese-Chinese SMT model, and improved the translation quality by 
converting Vietnamese syllables into Chinese characters with a pre-specified dictionary. 
Partially motivated by this work, we turn to Korean in terms of NMT models by fully exploiting the shared Sino-Korean vocabulary between Korean and Chinese.

\section{Sino-Korean Words and Chinese Characters}
\label{sect:Sino_Korean}
Korea belongs to the Chinese cultural sphere, which means that China has historically influenced regions and countries of East Asia. Before the creation of Hangul ({\em Korean alphabet}), all documents were written in Chinese characters, and Chinese characters were used continuously even after the creation of Hangul. 

Today, the standard writing system in Korea is Hangul, and the use of Chinese characters in Korean sentences is rare, but Chinese characters have left a significant influence on Korean vocabulary. 
About 290,000 (57\%) out of the 510,000 words in the {\em Standard Korean Language Dictionary} published by the {\em National Institute of Korean Language} belongs to Sino-Korean words, which were originally written in Chinese characters. Some Sino-Korean words do not currently have  corresponding Chinese words and their meanings and usage have changed in the process of introduction, but most of them have corresponding Chinese words. In Korean, Sino-Korean words are mainly used as literary or technical vocabulary and are often used in abstraction concepts and technical terms. The names of people and Korea place are mostly composed of Chinese characters, and newspapers and professional books occasionally use both Hangul and Chinese characters to clarify the meaning. Table~\ref{2-table} shows some news headlines that contain Chinese characters from the Korean news.

\begin{table}

\begin{center}
\begin{tabular}{m{7cm}}
\hline 
\begin{CJK*}{UTF8}{bsmi}\underline{北}\end{CJK*}\begin{CJK}{UTF8}{mj} 선전매체 “\end{CJK}\begin{CJK*}{UTF8}{bsmi}\underline{北美}\end{CJK*}\begin{CJK}{UTF8}{mj}관계도  “\end{CJK}\begin{CJK*}{UTF8}{bsmi}\underline{南北}\end{CJK*}\begin{CJK}{UTF8}{mj}관계처럼 대전환”\end{CJK} \\
\begin{CJK}{UTF8}{mj}3.1운동 100주년 맞아 장병 어깨에 원색( \end{CJK}\begin{CJK*}{UTF8}{bsmi}\underline{原色}\end{CJK*}\begin{CJK}{UTF8}{mj}) 태극기 부착\end{CJK} \\
\hline 
\end{tabular}
\end{center}
\caption{News headlines with Chinese characters. The underline denotes Chinese characters.}\label{2-table}
\end{table}

Since Korean belongs to alphabetic writing systems and is a language that does not have tones like Chinese, 
many homophones were created in their vocabulary in the process of translating the Chinese words into their language.
Around 35\% of the Sino-Korean words registered in the {\em Standard Korean Language Dictionary} belong to homophones.
Thus converting Sino-Korean words into (usually different) Chinese characters will have a similar impact as semantic disambiguation. For example, the Korean word uisa (\begin{CJK}{UTF8}{mj}의사\end{CJK} in Korean) has many homophones and can have several meanings. 
To clarify the meaning of the word uisa in Korean context, these words are occasionally written in Chinese characters as follows: \begin{CJK*}{UTF8}{gbsn}医师\end{CJK*} ({\em doctor}), \begin{CJK*}{UTF8}{gbsn}意思\end{CJK*} ({\em mind}), \begin{CJK*}{UTF8}{gbsn}义士\end{CJK*} ({\em martyr}), \begin{CJK*}{UTF8}{gbsn}议事\end{CJK*} ({\em proceedings}).

In addition, There is a difference between Chinese characters (Hanja) used in Korea and Chinese characters used in China. Chinese can be divided into two categories: Traditional Chinese and Simplified Chinese. Chinese characters used in China and Korea are Simplified Chinese and Traditional Chinese, respectively.

\section{The Proposed Approach}

The proposed approach for Korean-to-Chinese MT has two phases: Hangul-Hanja conversion and NMT model training. We first convert the Sino-Korean words of the Korean input sentences into Chinese characters, and convert the Traditional Chinese characters of the converted Korean input sentences into Simplified Chinese characters to share the common units between source and target vocabulary. Then we train NMT models with the converted Korean sentences as source data and the original Chinese sentences as target data.

For Hangul-Hanja conversion, we use open toolkit Hanjaro that is provided by the {\em Institute of Traditional Culture}\footnote{https://hanjaro.juntong.or.kr}. The Hanjaro can accurately convert Sino-Korean words into Chinese characters and is based on open toolkit UTagger (Shin and Ock \shortcite{Shin:12} in Korean) developed by the {\em Korean Language Processing Laboratory} of {\em Ulsan University}. More specifically, the Hanjaro first obtains tagging information about morpheme, parts of speech(POS) and homophones of a Korean sentence through the Utagger, and converts Sino-Korean words into corresponding Chinese characters by using this tagging information and pre-built dictionary. 
The UTagger is the Korean morphological tagging model 
which has a recall of 99.05\% on morpheme analysis and 96.76\% accuracy on POS and homophone tagging. Nguyen et al. \shortcite{Nguyen:19} significantly improved the performance Korean-Vietnamese NMT system by building a lexical semantic network for the special characteristics of Korean, which is using a knowledge base of the UTagger, and applying the Utagger to Korean tokenization.


For MT modeling, we use two types of NMT models: RNN based NMT and Transformer NMT models. We train the NMT models on parallel corpus processed through the Hangul-Hanja conversion above. 

\section{Experiments}

There have been many studies on how to segment Korean and Chinese text \cite{Zhao:08,Zhaob:08,Zhaob:13,Cai and Zhao:16,Deng:17}.
To find out which segmentation method has the highest translation performance,
we tried multiple segmentation strategies such as byte-pair-encoding \cite{Sennrichb:16}, jieba\footnote{https://pypi.org/project/jieba/} 
, KoNLP\footnote{http://konlpy.org} 
and so on. Eventually, we found that character-based segmentation for both languages can give the best performance.
Therefore, both Korean and Chinese sentences are segmented into characters for our NMT models.

\begin{table}
\begin{center}
\begin{tabular}{c|c|c|c}
\hline \textbf{Domains} & \textbf{Train} & \textbf{Validation} & \textbf{Test} \\ 
\hline
Society&	67363&	2,000&	2,000 \\
All	&258386	&5,000&	5,000 \\
\hline
\end{tabular}
\end{center}
\caption{The statistics for the parallel corpus extracted from Dong-A newspaper (The number of sentences). }\label{3-table} 
\end{table}

\subsection{Parallel Corpus}


We use two parallel corpora in our experiment. The first corpus is a Chinese-Korean parallel corpus of casual conversation and provided by {\em Semantic Web Research Center}\footnote{http://semanticweb.kaist.ac.kr} (SWRC). However, the SWRC corpus contained some incomplete data, so we removed the erroneous data manually. The parallel corpus consists of a set of 55,294 pairs of parallel sentences. 2,000 and 2,000 pairs from the parallel corpus were extracted as validation data and test data, respectively.

The second corpus (Dong-A) is collected from the online Dong-A newspaper\footnote{http://www.donga.com/ (Korean) and http://chinese.donga.com/ (Chinese)} by us. 
We collected articles on four domains, Economy (81,278 sentences), Society (71,363), Global (68,073) and Politics (61,208), to build two corpora as shown in 
Table~\ref{3-table}.

Since the sentences in the Dong-A newspaper are relatively long,  the maximum sequence length that we used to train the NMT model is set to 200. On the other hand, the maximum sequence length for SWRC corpus is set to 50 because each sentence in the SWRC corpus is short.


\subsection{NMT Models}

The Torch-based toolkit OpenNMT \cite{Klein:18} is used to build our NMT models, either RNN-based or Transformer. 

As for RNN-based models, we further consider two types of them, one with unidirectional LSTM encoder (uni-RNN) and the other with bidirectional LSTM based encoder (bi-RNN). For both
RNN based models, we use 2-layer LSTM with 500 hidden units on both encoder and decoder and use the global attention mechanism as described in \cite{Luong:15}. We use stochastic gradient descent (SGD) optimizer with the initial learning rate 1 and with decay rate 0.5. Mini-batch size is set to 64, and the dropout rate is set to 0.3. 

For our Transformer model, both the encoder and decoder are composed of a stack of 6 uniform layers, each built of two sublayers as described in \cite{Vaswani:17}. The dimensionality of all input and output layers is set to 512, and that of Feed-Forward Networks (FFN) layers is set to 2048. We set the source and target tokens per batch to 4096. For optimization, we used Adam optimizer \cite{Kingma:14} with $\beta_{1}$= 0.9, $\beta_{2}$= 0.98 to tune model parameters, and the learning rate is set by the warm-up strategy with steps 
8,000 ,and it decreases proportionally as the model training progresses. 

All of the NMT models are trained for 100,000 steps and checked the performance on the validation set after every 5,000 training steps. And we save the models every 5,000 training steps and evaluate the models using traditional machine translation evaluation metric.


\subsection{Results}
\label{sect:Results}
We used the BLEU score \cite{Papineni:02} as our evaluation metric.
Tables~\ref{4-table} and \ref{5-table} show the experimental results for SWRC corpus and Dong-A corpus, respectively. All NMT models, trained with Korean sentences converted through Hangul-Hanja conversion as source sentences, improve the translation performance on all test sets in comparison to the NMT models for the original sentence pairs.
The absolute BLEU improvement is about 1.57 on average for SWRC corpus and 0.93 on average for Dong-A corpus when applied the Hangul-Hanja conversion, respectively.

\begin{table}[t]
\begin{center}
\begin{tabular}{c|c c}
\hline \textbf{Systems} &  \multicolumn{2}{c}{\textbf{BLEU Score (Test set)}} \\
&w/o HH-Conv.&w/ HH-Conv \\
\hline \hline
uni-RNN	&33.14&	34.44\\
bi-RNN	&35.31&	36.66\\
Transformer	&35.47&	37.84\\
\hline 
\end{tabular}
\end{center}
\caption{ Experimental results of SWRC corpus. The HH-Conv refers to Hangul-Hanja conversion function. }\label{4-table}
\end{table}

\begin{table}
\begin{center}
\begin{tabular}{c|c|c c}
\hline \textbf{Systems} & \textbf{Domains}& \multicolumn{2}{c}{\textbf{BLEU Score }} \\

	&	& w/o HH-c. &	w/ HH-c \\
\hline \hline

\multirow{2}{4em}{uni-RNN}&Society	&36.25	&37.58\\
	&All	&39.84&	40.70\\
	\hline
\multirow{2}{3.5em}{bi-RNN}	&Society&	39.08& 40.00	\\
	&All	&41.76	&42.81\\
	\hline
\multirow{2}{5em}{Transformer}&	Society	&39.34&40.55\\
	&All&	44.70&	44.88\\

\hline
\end{tabular}
\end{center}
\caption{Experimental results of Dong-A corpus.}\label{5-table} 
\end{table}

Our proposed method is to improve the translation performance of NMT models by converting only Sino-Korean words into corresponding Chinese characters in Korean sentences using the Hanjaro and sharing the source vocabulary and the target vocabulary. 

In the work, we do not convert the entire Korean sentence into Chinese characters using a pre-specified dictionary and maximum matching mechanism as described in \cite{Zhao:13}. Unlike Chinese, which does not use inflectional morphemes, Korean belongs to an agglutinative language that tends to have a high rate of affixes or morphemes per word. Since some Korean syllables do not have corresponding Chinese characters, so converting all Korean syllables of Korean sentence into Chinese characters is an impossible mission. In fact, we built a bilingual dictionary for Korean and Chinese and used maximum matching mechanism to convert all the affixes and inflectional morphemes of Korean sentences into Chinese characters and trained an RNN based NMT model, but the performance was even lower.

In our implementation, we estimate that the main reason for improving performance is to make the distinction between homophones clearer by converting Sino-Korean words into Chinese characters. Many of the Korean vocabularies that employ the alphabetical writing system are homophones, which can confuse meaning or context. Especially, as mentioned in Section~\ref{sect:Sino_Korean}, 35\% of Sino-Korean words are homophones. Therefore, it is possible to clarify the distinction between homophones by applying Hangul-Hanja conversion to Korean sentences, which leads to performance improvement in Korean-to-Chinese MT.

\section{Analysis}

\subsection{Analysis on Sino-Korean word Conversion}

In this subsection, we will analyze the conversion from Sino-Korean words to Chinese characters.
To estimate how much Chinese characters converted from Sino-Korean words by Hangul-Hanja conversion function are included in the corresponding reference sentence, we propose {\em ratio of including the same Chinese character between the converted Korean sentence and Chinese sentence (reference sentence)} (ROIC):

\begin{equation}
ROIC = \frac{\sum_{w_i} f(w_i)}{\mathopen|w\mathclose|}
\end{equation}
where \(|w|\) is the number of Chinese words in converted Korean sentence, \(f(w_i)\) is 1 if the Chinese word \(w_i\) of the converted Korean sentence is included in the corresponding Chinese sentence, and 0 otherwise. 
For example, in the second example of Table~\ref{1-table}, because the five Chinese words such as \begin{CJK*}{UTF8}{gbsn}两国\end{CJK*} ({\em two countries}), \begin{CJK*}{UTF8}{gbsn}领域\end{CJK*} ({\em area}), \begin{CJK*}{UTF8}{gbsn}共同\end{CJK*} ({\em common}), \begin{CJK*}{UTF8}{gbsn}利益\end{CJK*} ({\em interests}), \begin{CJK*}{UTF8}{gbsn}确认\end{CJK*} ({\em confirm}) are commonly observed between the converted Korean sentence and the reference sentence except for \begin{CJK*}{UTF8}{gbsn}广范\end{CJK*} ({\em abroad}), so we say that the ROIC of the converted Korean sentence is \(\frac{5}{6}\) (83.33\%).
We perform analysis of Sino-Korean word conversion in two separate ways: ROIC for Chinese word and ROIC for Chinese character.

\begin{figure}[t]
\centering
\includegraphics[width=8cm]{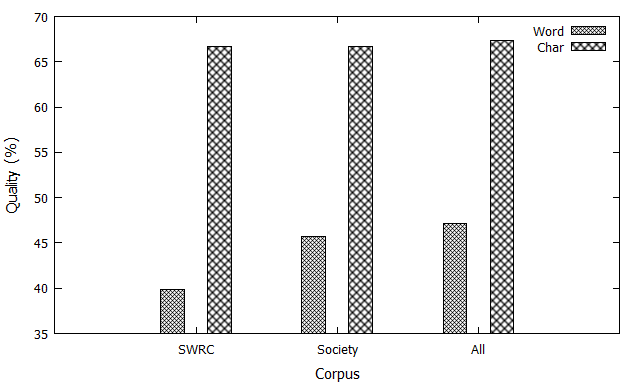}
\caption{ROIC of each corpus. Word and Char denote the ROIC for Chinese word and the ROIC for Chinese character, respectively.} \label{1-figure}
\end{figure}

Fig.~\ref{1-figure} presents the ROIC of each corpus. It can be observed that for each corpus, more than 40\% of the converted Chinese words or more than 65\% of the converted Chinese characters are included in the reference sentence. So we can see that source vocabulary and target vocabulary share many words after converting Sino-Korean words into Chinese characters. Sharing source vocabulary and target vocabulary is especially useful for same alphabet languages, or for domains where professional terms are written in English \cite{Zhanglon:18}. Therefore, we set to share the source vocabulary and the target vocabulary of our NMT models, which leads to performance improvement.

\begin{figure}[ht]
\centering
\includegraphics[width=8cm]{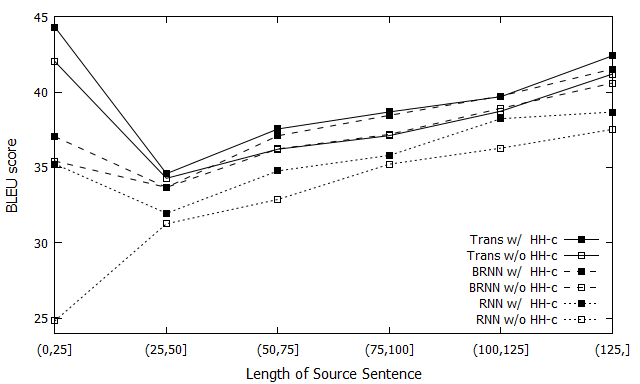}
\caption{BLEU scores for the translation of sentences with different lengths.}\label{2-figure}
\end{figure}

\subsection{Analysis of Translation Performance according to Different Sentence Lengths}

Following Bahdanau et al. \shortcite{Vaswani:17}, we group sentences of similar lengths together and compute BLEU scores, which are presented in Fig.~\ref{2-figure}. we conduct this analysis on Society corpus. It shows that our method leads to better translation performance for all the sentence lengths. Since we set the Maximum sentence length to 200 for the Society corpus, we also can see that the performance continues to improve when the length of the input sentence increases.

\begin{table*}[t!]
\begin{center}
\begin{tabular}{p{2.5cm}|p{12cm}}
\hline \textbf{Systems} & \textbf{Sentences} \\ \hline\hline
Korean & \begin{CJK}{UTF8}{mj}이 지역에 사는 \underline{유지}*들이 이 마을을 \underline{유지}**하고 관리해나가고 있다.\end{CJK} \\
HH-Convert & \begin{CJK}{UTF8}{mj}이 \end{CJK}{\begin{CJK*}{UTF8}{gbsn}地域\end{CJK*}}\begin{CJK}{UTF8}{mj}에 사는  \end{CJK}\begin{CJK*}{UTF8}{gbsn}\underline{有志}*\end{CJK*}\begin{CJK}{UTF8}{mj}들이 이 마을을 \end{CJK}\begin{CJK*}{UTF8}{gbsn}\underline{维持}**\end{CJK*}\begin{CJK}{UTF8}{mj}하고 \end{CJK}\begin{CJK*}{UTF8}{gbsn}管理\end{CJK*}\begin{CJK}{UTF8}{mj}해나가고 있다.\end{CJK}\\
Chinese & \begin{CJK*}{UTF8}{gbsn}在这个区域生活的\underline{有志之士}*在\underline{维护}**和管理这个小区。 \end{CJK*}  \\
English &  The \underline{community leaders}* living in this area are \underline{maintaining}** and managing this community. \\ 
Trans w/o HH-c & \begin{CJK*}{UTF8}{gbsn}居住在该地区的\underline{维持}**和管理村庄。 \end{CJK*}\\
Trans w/ HH-c & \begin{CJK*}{UTF8}{gbsn}居住在该地区的\underline{有志}*们\underline{维持}**这个村子，并进行管理。 \end{CJK*}\\

\hline \hline
Korean & \begin{CJK}{UTF8}{mj}\underline{이성}* 간의 교제는 \underline{이성}**에 따라 해야 한다.\end{CJK} \\
HH-Convert & \begin{CJK*}{UTF8}{gbsn}\underline{异性}* 间\end{CJK*}\begin{CJK}{UTF8}{mj}의  \end{CJK}\begin{CJK*}{UTF8}{gbsn}交际\end{CJK*}\begin{CJK}{UTF8}{mj}는  \end{CJK}\begin{CJK*}{UTF8}{gbsn}\underline{理性}**\end{CJK*}\begin{CJK}{UTF8}{mj}에 따라 해야 한다. \end{CJK}\\
Chinese & \begin{CJK*}{UTF8}{gbsn}\underline{异性}*之间交往应该保持\underline{理性}**。 \end{CJK*}  \\
English &  A romantic relationship between the \underline{opposite sex}* should be \underline{rational}**. \\
Trans w/o HH-c & \begin{CJK*}{UTF8}{gbsn}\underline{理性}**间的交往应遵从\underline{理性}**。 \end{CJK*}\\
Trans w/ HH-c & \begin{CJK*}{UTF8}{gbsn}\underline{异性}*之间的交往应该根据\underline{理性}**进行。 \end{CJK*}\\
\hline \hline
Korean & \begin{CJK}{UTF8}{mj}그는 천연\underline{자원}*을 탐사하는 임무에 \underline{자원}**했다.\end{CJK} \\
HH-Convert & \begin{CJK}{UTF8}{mj}그는  \end{CJK}\begin{CJK*}{UTF8}{gbsn}天然\underline{资源}*\end{CJK*}\begin{CJK}{UTF8}{mj}을  \end{CJK}\begin{CJK*}{UTF8}{gbsn}探查\end{CJK*}\begin{CJK}{UTF8}{mj}하는  \end{CJK}\begin{CJK*}{UTF8}{gbsn}任务\end{CJK*}\begin{CJK}{UTF8}{mj}에  \end{CJK}\begin{CJK*}{UTF8}{gbsn}\underline{自愿}**\end{CJK*}\begin{CJK}{UTF8}{mj}했다.\end{CJK}\\
Chinese & \begin{CJK*}{UTF8}{gbsn}他\underline{自愿}**参加勘探自然\underline{资源}**的任务。 \end{CJK*}  \\
English &  He \underline{volunteered}** for the task of exploring natural \underline{resources}*. \\
Trans w/o HH-c & \begin{CJK*}{UTF8}{gbsn}他为探测天然\underline{资源}**的任务提供了\underline{资源}**。 \end{CJK*}\\
Trans w/ HH-c & \begin{CJK*}{UTF8}{gbsn}他\underline{自愿}**担任探测天然\underline{资源}*的任务。 \end{CJK*}\\

\hline \hline
Korean & \begin{CJK}{UTF8}{mj}\underline{의사}*의 꿈은 포기했지만, 가족들은 그의 \underline{의사}**를 존중해주었다.\end{CJK} \\
HH-Convert & \begin{CJK*}{UTF8}{gbsn}\underline{医师}*\end{CJK*}\begin{CJK}{UTF8}{mj}의 꿈은   \end{CJK}\begin{CJK*}{UTF8}{gbsn}抛弃\end{CJK*}\begin{CJK}{UTF8}{mj}했지만,  \end{CJK}\begin{CJK*}{UTF8}{gbsn}家族\end{CJK*}\begin{CJK}{UTF8}{mj}들은 그의   \end{CJK}\begin{CJK*}{UTF8}{gbsn}\underline{意思}**\end{CJK*}\begin{CJK}{UTF8}{mj}를 \end{CJK}\begin{CJK*}{UTF8}{gbsn}尊重\end{CJK*}\begin{CJK}{UTF8}{mj}해주었다. \end{CJK}\\
Chinese & \begin{CJK*}{UTF8}{gbsn}虽然放弃了\underline{医生}*的梦想,但家人也尊重他的\underline{意愿}**。 \end{CJK*}  \\
English & Although he gave up on his dream of becoming a \underline{doctor}*, his family respected his \underline{wishes}**. \\
Trans w/o HH-c & \begin{CJK*}{UTF8}{gbsn}虽然\underline{医生}*的梦想放弃了，但是家人却尊重了他的意向。 \end{CJK*}\\
Trans w/ HH-c & \begin{CJK*}{UTF8}{gbsn}虽然放弃了\underline{医生}*的梦想，但家人却尊重了他的\underline{意愿}**。 \end{CJK*}\\

\hline \hline
\end{tabular}
\end{center}
\caption{\label{6-table} Translation results of sentences with two homophones. The HH-Convert is Korean sentence converted by Hangul-Hanja conversion of the Hanjaro. Trans w/o HH-c and Trans w/ HH-c are the translation results of Transformer baseline model and Transformer using our method, respectively. The underline denotes homophone and the number of stars(*) distinguishes the meanings of the homophone in each example. In Chinese, English, and translation results, they denote words that are equivalent to the homophones in the sense of meaning.}
\end{table*}

\subsection{Analysis of Homophones Translation}
In this subsection, we translate several sentences that contain two homophones and analyze how the Sino-Korean word conversion makes the distinction between homophones more apparent. 
We translated the sentences using the Transformer model trained with the Dong-A corpus.
Table~\ref{6-table} presents the translation results of sentences with two homophones.

We can see that our NMT model clearly distinguishes between homophones for all examples, but the baseline model does not distinguish or translate homophones. 
For example, in the first example, the baseline model does not translate \begin{CJK}{UTF8}{mj}유지*\end{CJK} ({\em community leader}). In the second and third example, the baseline model translated them into the same words without distinguishing between the homophones. 
In the last example, \begin{CJK}{UTF8}{mj}의사\end{CJK}** ({\em wishes}) was improperly translated into \begin{CJK*}{UTF8}{gbsn}意向\end{CJK*} ({\em intention}).
Therefore, as mentioned in Section~\ref{sect:Results}, these results indicate that our method helps distinguish homophones in Korean-to-Chinese machine translation.

\section{Conclusion}

This paper presents a simple novel method exploiting the shared vocabulary of a low-resource language pair for better machine translation. In detail, we convert Sino-Korean words in Korean sentences into Chinese characters and then train machine translation model with the converted Korean sentences as source sentences.
Our proposed improvement has been verified effective over RNN-based and latest Transformer NMT models. Besides, we regard that this is the first attempt which takes a linguistically motivated solution for low-resource translation using NMT models. Although this proposed method seems only suitable for the language pair of Korean and Chinese, it has enormous potential to work for any language pair which shares a considerable vocabulary from their shared history. 

\end{document}